\pdfoutput=1

\documentclass[11pt]{article}

\usepackage[final]{acl}

\usepackage{times}
\usepackage{latexsym}

\usepackage[T1]{fontenc}

\usepackage[utf8]{inputenc}

\usepackage{microtype}

\usepackage{inconsolata}

\usepackage{algorithm}
\usepackage{algorithmic}
\usepackage{graphicx}
\usepackage{booktabs}
\usepackage{tabularx}
\usepackage{multirow}
\usepackage{adjustbox}
\usepackage{url}
\usepackage{amsmath}
\usepackage{xcolor}
\usepackage{bbding}
\usepackage{amssymb}
\title{Intuitive Fine-Tuning: \\ Towards Simplifying Alignment into a Single Process}

\author{Ermo Hua$^1$, Biqing Qi$^{2,*}$, Kaiyan Zhang$^{1}$, \\
        \textbf{Kai Tian$^{1}$, Xingtai Lv$^{1}$, Ning Ding$^{1}$, Bowen Zhou$^{1,2,}$\thanks{Corresponding Author}} \\
        $^1$Department of Electronic Engineering, Tsinghua University, Beijing, China \\
        $^2$Shanghai AI Laboratory, Shanghai, China \\
        \texttt{hem23@mails.tsinghua.edu.cn}\\ \texttt{qibiqing@pjlab.org.cn}, 
        \texttt{zhoubowen@tsinghua.edu.cn}}

\begin{document}
\maketitle
\begin{abstract}
Supervised Fine-Tuning (SFT) and Preference Optimization (PO) are key processes for aligning Language Models (LMs) with human preferences post pre-training.
While SFT excels in efficiency and PO in effectiveness, they are often combined sequentially without integrating their optimization objectives. 
This approach ignores the opportunities to bridge their paradigm gap and take the strengths from both.
In this paper, we interpret SFT and PO with two sub-processes — \textit{Preference Estimation} and \textit{Transition Optimization} — defined at token level within the Markov Decision Process (MDP). This modeling shows that SFT is only a special case of PO with inferior estimation and optimization.
PO estimates the model's preference by its entire generation, while SFT only scores model's subsequent predicted tokens based on prior tokens from ground truth answer. These priors deviates from model's distribution, hindering the preference estimation and transition optimization.
Building on this view, we introduce \textit{\textbf{Intuitive Fine-Tuning (IFT)}} to integrate SFT and PO into a single process. Through a temporal residual connection, IFT brings better estimation and optimization by capturing LMs' intuitive sense of its entire answers. But it solely relies on a single policy and the same volume of non-preference-labeled data as SFT.
Our experiments show that IFT performs comparably or even superiorly to SFT and some typical PO methods across several tasks, particularly those requires generation, reasoning, and fact-following abilities. An explainable Frozen Lake game further validates the effectiveness of IFT for getting competitive policy.
Code is available at \textcolor{blue}{\url{https://github.com/TsinghuaC3I/Intuitive-Fine-Tuning}}.
\end{abstract}

\section{Introduction}
\label{sec:introduction}
Large Language Models (LLMs) have demonstrated remarkable powerful potential across various downstream tasks after pre-training on large-scale corpora \citep{brown2020language, achiam2023gpt, zhou2024generative}.
However, their instruction-following skills and trustworthiness still fall short of expectations \citep{bender2021dangers, bommasani2021opportunities, li2022trust}.
Therefore, algorithms such as Supervised Fine-Tuning (SFT) and Reinforcement Learning from Human Feedback (RLHF) \citep{ziegler2019fine, ouyang2022training, lee2023rlaif} are used to further enhance LLMs' abilities and align them better with human preferences. 

Considering the limited effectiveness of SFT and the high cost of data construction and training computation for RLHF, these two methods are often combined to leverage their respective strengths. Unfortunately, they are typically implemented as a sequential recipe constrained by the paradigm gap between SFT and early RLHF methods, stemming from differences in loss functions, data formats, and the requirement for auxiliary models. 

\begin{figure}[t]
    \includegraphics[width=\columnwidth]{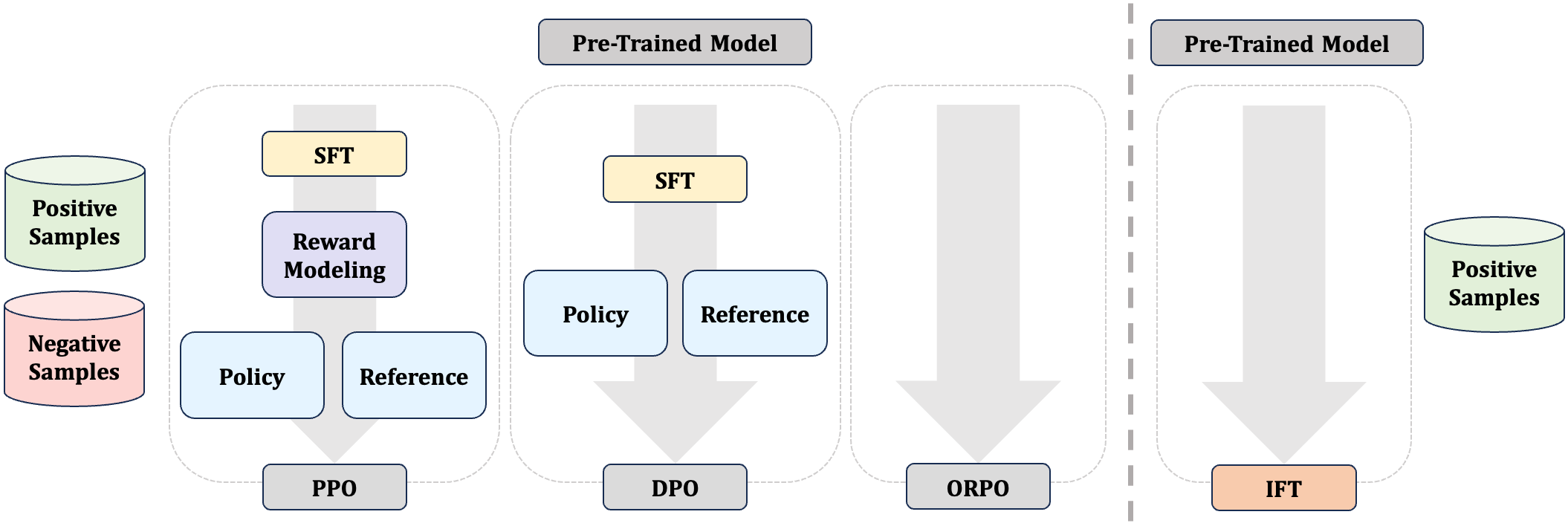}
    \caption{Comparison of Alignment Methods. IFT conducts alignment solely relying on positive samples and a single policy, starting from a pre-trained base model. IFT shows similar efficiency as SFT and effectiveness as PO methods.}
    \label{fig:framework}
\end{figure}


Recently, a method named Direct Preference Optimization (DPO) \citep{rafailov2024direct} was proposed to integrate Reward Modeling and Policy Optimization into one single procedure using a loss function derived from Proximal Policy Optimization (PPO) \citep{schulman2017proximal}. This approach demonstrates the potential to unify SFT and RLHF for the first time.
Henceforth, many extended methods have been tried to realize this objective by bridging the gap between SFT and DPO. Some of them \citep{ethayarajh2024kto, hong2024reference, zhang2024negative} aim to transform the contrastive loss of DPO into a SFT-like cross-entropy loss, learning positive samples similar to SFT while unlearning negative samples resort to Unlikelihood Training \citep{welleck2019neural}. Some others get rid of the preference-labeling process before training, switching to collect samples and labels/rewards in an online manner \citep{liu2023statistical, yuan2024self, guo2024direct, calandriello2024human, tajwar2024preference}, or just treating the SFT targets and online policy generations as positive and negative samples respectively \citep{xiong2023iterative, chen2024self, mitra2024orca, liu2024extensive}.
Nevertheless, preference-labeled pairwise data is still essential, and the need for reference model only becomes unnecessary in some cases. Thus the core differences between SFT and Preference Optimization (PO) are not eliminated thoroughly. To address this challenging issue, a deeper and more unified understanding of them are needed.

In this paper, we attempt to explain the similarities and differences between SFT and PO by defining Preference Estimation and Transition Optimization in terms of state-action pairs within the Markov Decision Process (MDP) framework. Through this modeling, we demonstrate that SFT is simply a specialized case of PO with inferior estimation and optimization than other methods.
To estimate the policy preference, PO collects sentence-level negative samples from policy for each initial instruction. However, SFT only samples subsequent token for each intermediate state of ground truth answer, which leads to a biased estimation of policy preference and an inferior alignment performance. 

Depending on this understanding, we introduce a unified alignment algorithm named Intuitive Fine-Tuning (IFT). Drawing inspiration from the human ability to grasp a intuitive sense of an answer after hearing a question, IFT employs a Temporary Residual Connection across tokens to approximate policy’s entire answer for each instruction. This approach helps IFT better estimate the policy's preference than SFT, achieving alignment performance comparable or even superior to the sequential recipe of SFT and Preference Optimization. Additionally, IFT requires only a single policy model, and the same volume and format of data as SFT, enjoying both data and computation efficiency.
These characteristics of IFT are advantageous in domains where preference data is unavailable or expensive to collect.

Our main contribution are three folds: 

\textbf{(1)} Through defining Preference Estimation and Transition Optimization using the MDP, we demonstrate that SFT is only a special case of Preference Optimization. The similarities and differences of SFT, PPO and online/offline DPO are also compared within this framework; 

\textbf{(2)} We introduce Intuitive Fine-tuning (IFT), a deeply unified version of SFT and Preference Optimization. It utilizes temporary residual connections to extract the model's generation preference given the initial instructions. IFT enjoys the similar efficiency as SFT on negative sampling, but can better estimate and optimize the policy preference.

\textbf{(3)} Through experiments on several benchmarks, we validate that IFT performs comparably or superiorly to SFT and various Preference Optimization methods. An explainable toy-setting Frozen Lake further demonstrates the effectiveness of IFT.

\begin{figure*}[t]
    \centering
    \includegraphics[width=\textwidth]{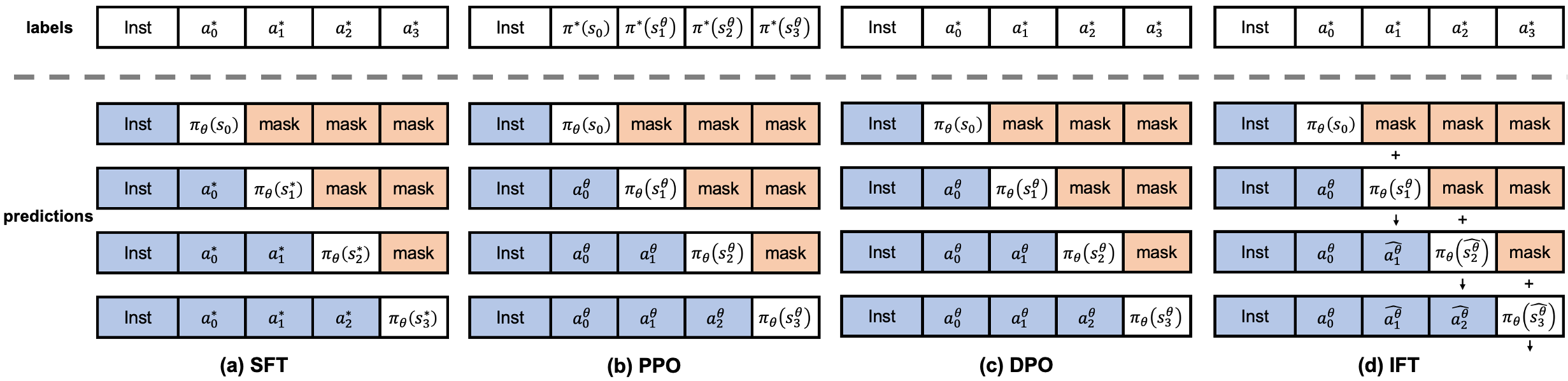}
    \caption{The Training Paradigm of Different Methods. Symbol $*$ and $\theta$ denote human and model respectively, with $a_i^*=\pi^*(s_i^*)$ and $s_{i+1}^*=[s_i^*, a_i^*]$, similarly for $\theta$. SFT uses priors deviating from model distribution, resulting in a more biased estimation of model preferences compared to PPO and DPO. IFT achieves a better estimation than SFT by Temporary Residual Connections across tokens. This approach passes the residual embedding from one token to the next, creating a more accurate prior while maintaining the data and computational efficiency of SFT.}
    \label{fig:formulation}
\end{figure*}

\section{Preliminaries}
\label{sec:preliminaries}
\subsection{MDP in Language Models}

The MDP applied to LMs can be formally described as a tuple $\mathcal{M}=(\mathcal{S}, \mathcal{A}, \mathcal{T}, r, \rho_0)$, where $\mathcal{S}$ is the state space comprising ordered permutations of vocabularies, $\mathcal{A}$ is the action space consisting of vocabularies defined by the tokenizer, $\mathcal{T}$ is the transition matrix indicating token generation probabilities for given states, $r$ represents rewards for state-action pairs, and $\rho_0$ is the initial state typically based on given instructions. See more details in Appendix A.1.

The primary objective of Language Modeling is to train a policy $\pi_\theta$ with $\mathcal{T}_\theta$ to mimic a human policy $\pi^*$ with $\mathcal{T}^*$, aiming for the two transition matrices to become identical:
\begin{equation}
    \forall s \in \mathcal{S}, a \in \mathcal{A}: \mathcal{T}_\theta(a|s) \rightarrow \mathcal{T}^*(a|s)
\end{equation}

This process can also be expressed using another state-state transition matrix $T$:
\begin{equation}
    \forall~ s,s' \in S: T_\theta(s'|s) \rightarrow T^*(s'|s)
\end{equation}
where $T$ is equivalence to $\mathcal{T}$, but instead, indicating the transition probability between states.

\subsection{Preference Estimation}

We define the preference $\mathcal{P}$ of policy $\pi$ given an initial instruction $\rho_0$ as a mapping:
\begin{equation}
    \mathcal{P}(\rho_0): \rho_0 \rightarrow [\pi(\rho_0), \pi(s_1), \pi(s_2), ...]
\end{equation}
where $s_{i+1}=[s_i, a_i]$ , $a_i=\pi(s_i)$ and $s_0=\rho_0$. 

During alignment, the model preference gradually approaches the human preference:
\begin{equation}
    \mathcal{P_\theta}(\rho_0) \rightarrow \mathcal{P^*}(\rho_0)
\end{equation}
\begin{equation}
    \begin{aligned}
        & \mathcal{P_\theta}(\rho_0): \rho_0 \rightarrow [\pi_\theta(\rho_0), \pi_\theta(s_1^\theta), \pi_\theta(s_2^\theta), \ldots] \\
        & \mathcal{P^*}(\rho_0): \rho_0 \rightarrow [\pi^*(\rho_0), \pi^*(s_1^*), \pi^*(s_2^*), \ldots]
    \end{aligned}
\end{equation}
As the truly preferences are difficult to obtain, alignment is usually conducted based on the Preference Estimation of model and human, denoted as $\hat{\mathcal{P}_\theta}$ and $\mathcal{\hat{P^*}}$ respectively. The estimations from some typical methods are listed in Table \ref{tab:formular}.

To make preference optimizable, the policy's preference can also be expressed as follows:
\begin{equation}
    \mathcal{P}(\rho_0) = \{\mathcal{T}(a|s) | \forall a \in \mathcal{A}, s \in \mathcal{S}_{\rho_0}\}
\end{equation}
Here, $\mathcal{S}_{\rho_0}$ denotes a conditional state space that constrained by the initial state $\rho_0$, within which each state can only be initially derived from $\rho_0$.
Consequently, the model preference can be optimized through transition matrix, named Transition Optimization.

\subsection{Transition Optimization}

Ideally, we want to align the state-action transition matrix between model and human in a $\rho_0$-constrained state space:
\begin{equation}
    \forall a \in \mathcal{A}, s \in \mathcal{S}_{\rho_0}: \mathcal{T}_\theta(a, s) \rightarrow \mathcal{T^*}(a, s)
\end{equation}
which is equivalent to the following format expressed by state-state transition matrix:
\begin{equation}
    \forall s \in \mathcal{S}_{\rho_0}: T_\theta(s, \rho_0) \rightarrow T^*(s, \rho_0)
\end{equation}

However, considering the limited data, only matrix elements representing state-action/state-state pairs contained in the dataset $\mathcal{D}$ would be aligned. Given a data sample with instruction $\rho_0$ and target answer with length-$N$, the objective would be $\forall a \in \mathcal{A}, n \in [0, N], \rho_0\in\mathcal{D}, s_n^*\in\mathcal{S}_{\rho_0}^*$:
\begin{equation}
    \mathcal{T}_\theta(a, s_n^*) \rightarrow \mathcal{T^*}(a, s_n^*)
\end{equation}
Or equivalent to $\forall n \in [0, N], \rho_0\in\mathcal{D}, s_n^*\in\mathcal{S}_{\rho_0}^*$:
\begin{equation}
    T_\theta(s_n^*, \rho_0) \rightarrow T^*(s_n^*, \rho_0)
\end{equation}
where 
$s_0^*=\rho_0$, $T^*(\rho_0|\rho_0)=T_\theta(\rho_0|\rho_0)=1$, and $s_i^*$ denotes the intermediate state of target answer.

Consequently, the loss function can be derived from the disparities of the transition matrices between model and human. Some typical loss function are listed in Appendix A.4.


\section{From SFT to Preference Optimization}
We reformulate SFT, PPO and DPO using the aforementioned framework, detailed in Table \ref{tab:formular} and Appendix A.4. A more comprehensible version is presented in Figure \ref{fig:formulation}. To compare the differences between them, we begin by introducing a fundamental theorem and corollary:

\textbf{Theorem} \textit{Given a set of events $\mathcal{Z}$, the probability of any event $z \in \mathcal{Z}$ is between 0 and 1, i.e., $\forall z \in \mathcal{Z}: 0 \leq P(z) \leq 1$. If all events are mutually independent, the sum of their probabilities equals 1, i.e., $1 = \sum_{z \in \mathcal{Z}}P(z)$. The event $z^*$ with the highest probability has a probability greater than or equal to any other event, i.e., $\forall z \in \mathcal{Z}: 0 \leq P(z) \leq P(z^*) \leq 1$.}


\textbf{Corollary} \textit{LMs consistently assign higher probabilities to their own greedy predictions than to human preference:}
\begin{equation}
    \forall s \in S: \mathcal{T}_\theta(\pi^*(s), s) \leq \mathcal{T}_\theta(\pi_\theta(s), s) \leq 1
\end{equation}
\textit{thus LMs tend to assign higher probabilities to its own generation than to target answer given the same initial instruction $\forall n \in [0,N], s_n^*\in\mathcal{S}_{\rho_0}^*, s_n^\theta\in\mathcal{S}_{\rho_0}^\theta$:}
\begin{equation}
    T_\theta(s_{n}^*, \rho_0) \leq T_\theta(s_{n}^\theta, \rho_0) \leq 1
\end{equation}
\textit{where $N$ represents the length when the generation reaches the EOS token or the truncation length.}

\textbf{SFT} provides an unbiased estimation of human preference, but a biased estimation for model:
\begin{equation}
    \hat{\mathcal{P_\theta}}(\rho_0): \rho_0 \rightarrow [\pi_\theta(\rho_0), \pi_\theta(s_1^*), \pi_\theta(s_2^*), \ldots]
\end{equation}
which is caused by wrong prior state when predicting each subsequent token. Consequently, the Transition Optimization objective of SFT:
\begin{equation}
    T_\theta(s_{n}^*, s_{n-1}^*) \rightarrow T^*(s_{n}^*, s_{n-1}^*)
\end{equation}
secretly sets $T_\theta(s_{n-1}^*, \rho_0)=1$ during aligning $T_\theta(s_{n}^*, \rho_0)$ with $T^*(s_{n}^*, \rho_0)$. This makes an overestimation of the transition probabilities and preference of model, leading to an inferior optimization progress in SFT. Thus Preference Optimization is needed for further preference alignment.

\textbf{PPO} shows an unbiased estimation of model preference, while employing a progressively unbiased estimation of human preference:
\begin{equation}
    \hat{\mathcal{P^*}}(\rho_0): \rho_0 \rightarrow [\pi^*(\rho_0), \pi^*(s_1^\theta), \pi^*(s_2^\theta), \ldots]
\end{equation}
Initially biased, this estimation gradually becomes unbiased as the model aligns with human preference over time. As $T_\theta(s_n^\theta, \rho_0)$ is consistently closer to 1 than $T_\theta(s_{n-1}^*, \rho_0)$, PPO provides an closer approximation than SFT to the actual circumstances of model in Transition Optimization:
\begin{equation}
    T_\theta(\hat{s_{n}^*}, s_{n-1}) \rightarrow T^*(\hat{s_{n}^*}, s_{n-1})
\end{equation}
which sets $T_\theta(s_n^\theta, \rho_0)=1$ and $\hat{s_n^*}=\pi^*(s_{n-1}^\theta)$. However, estimating $\pi^*(s_{n-1}^\theta)$ is at the expense of preference-labeling, reward modeling and online sampling. 

\textbf{DPO} theoretically achieves the best estimation across all scenarios, even without reward modeling. However, obtaining pairwise preference data online is costly, as it requires real-time negative sampling from model and preference labeling by human. Thus, mainstream implementations often rely on off-policy  negative samples out-of-distribution from the optimized model, which may yield unstable and sub-optimal results due to biased preference estimation and inferior transition optimization.

\begin{table}[t]
    \centering
    \begin{adjustbox}{width=\columnwidth}
        \begin{tabular}{ccccc}
            \toprule
            \multicolumn{2}{c}{\multirow{2}{*}{\textbf{Method}}} & \multicolumn{2}{c}{\textbf{Preference Estimation}} & \multirow{2}{*}{\textbf{Transition Optimization}} \\
            \cmidrule(lr){3-4}
            & & $\hat{s_n^*}$ in $\hat{\mathcal{P^*}}$ & $\hat{s_n^\theta}$ in $\hat{\mathcal{P_\theta}}$ & \\
            \midrule
            \multicolumn{2}{c}{Truly} & $s_n^*$ & $s_n^\theta$ & $T_\theta(s_n^*, \rho_0) \rightarrow T^*(s_n^*, \rho_0)$ \\
            \midrule
            \multicolumn{2}{c}{SFT} & $s_n^*$ & $s_n^*$ & $T_\theta(s_{n}^*, s_{n-1}^*) \rightarrow T^*(s_{n}^*, s_{n-1}^*)$ \\
            \midrule
            \multicolumn{2}{c}{PPO} & $s_n^\theta$ & $s_n^\theta$ & $T_\theta(\hat{s_{n}^*}, s_{n-1}^\theta) \rightarrow T^*(\hat{s_{n}^*}, s_{n-1}^\theta)$ \\
            \midrule
            \multirow{2}{*}{DPO} & online & $s_n^*$ & $s_n^\theta$ & $T_\theta(s_n^*, \rho_0) \rightarrow T^*(s_n^*, \rho_0)$ \\
            \cmidrule(lr){2-5}
            & offline & $s_n^*$ & $s_n^{\theta^-}$ & $\hat{T_\theta}(s_n^*, \rho_0) \rightarrow T^*(s_n^*, \rho_0)$ \\
            
            \bottomrule 
        \end{tabular}
    \end{adjustbox}
    \caption{Reformulation of SFT, PPO and DPO}
    \label{tab:formular}
\end{table}


\section{Method}
\label{sec:method}
While SFT is data and computation-efficient, it has an inferior approximation for both Preference Estimation and Transition Optimization. On the other side, Preference Optimization (represented by PPO and DPO) enjoys better approximation at the expense of preference data construction. We hope to make good use of their strength, using solely target data as SFT but having a similar approximation as Preference Optimization. See pseudo code in Appendix B.3.

\subsection{Intuitive Preference Estimation}

A key distinction between SFT and Preference Optimization is whether the full distribution of model's preference for each initial instruction is sampled. Preference Optimization samples the policy's entire answer to estimate its preference, ensuring each generation relies on the prior adheres to the model’s distribution. But SFT only samples subsequent tokens the intermediate state of the target answer, the used prior may be far away from the model preference, leading to inferior preference estimation for model.

To obtain a prior state estimation $\hat{s_i^\theta}$ closer to model distribution, we introduce a model-based distribution disturbance function $\delta_\theta$ for the biased prior state:
\begin{equation}
    \hat{s_i^\theta} = \delta_\theta(s_i^*) \\ = (1-\lambda)s_i^* + \lambda\pi_\theta(s_{i-1}^*)
\end{equation}
which can also be interpreted as a temporal residual connection that passes the residual embedding from one token to the next. Through this approach, model can predict not only the next token from intermediate state of target answer, but also develop an intuitive sense to the entire answer generation solely based on the initial instruction, deriving more unbiased prior and accurate Preference Estimation for model:
\begin{equation}
    \hat{\mathcal{P_\theta}}(\rho_0) = [(1-\lambda)\mathcal{P_\theta}^{sft} + \lambda\mathcal{P_\theta}^{truly}](\rho_0)
\end{equation}
With improved Preference Estimation, we achieve a Transition Optimization process closer to the original objective $\forall n \in [0, N], \rho_0\in\mathcal{D}, s_n^*\in\mathcal{S}_{\rho_0}^*$:
\begin{equation}
    \hat{T_\theta}(s_n^*, \rho_0) \rightarrow T^*(s_n^*, \rho_0)
\end{equation}
where $s_0^*=\rho_0$ and $\hat{T_\theta}(s_n^*, \rho_0) = \prod\limits_{i=0}^{n-1}T_\theta(s_{i+1}^*, \hat{s_i^\theta})$. 

This objective can be optimized by the following loss function, which quantifies the disparities of transition between model and human:
\begin{equation}
    \mathcal{L}(\mathcal{T}_\theta,\delta_\theta) 
    = \mathbb{E}
    \left[-\sum\limits_{n=0}^N\log\mathcal{T}_\theta(a_i^*, \delta_\theta(s_i^*))\right]
\end{equation}
where $a_i^*=\pi^*(\delta^*(s_i^*))=\pi^*(s_i^*)$. See Appendix A.2 for complete derivation.

\subsection{Dynamic Relation Propagation}

The Intuitive Preference Estimation implicitly performs Dynamic Relation Propagation, during which the generation of future tokens will be influenced by the prediction accuracy of current token. 

However, limited by the parallel computing mode, the gradient map could only be built on the same time-step. Thus, the current generated tokens is unable to obtain gradient feedback from the future generated tokens. Therefore, we reformulate the loss function by a differentiable cumulative-summation to get around this limitation:
\begin{equation}
    \mathcal{L}_{\text{IFT}}
    = \mathbb{E}
    \left[-\sum\limits_{n=0}^N\sum\limits_{i=n}^N\log\mathcal{T}_\theta(a_i^*, \delta_\theta(s_i^*))\right]
\end{equation}

This reformulation implicitly satisfies the Bellman Equation for each state, which guarantees the optimization enjoys both of the effectiveness as RLHF and efficiency as SFT:
\begin{equation}
    V_\theta(\hat{s_n^\theta}) = \exp\bigg(-\mathcal{L}\big(\hat{T_\theta}(s_n^*, \rho_0)\big)\bigg)
\end{equation}
The derivation is in Appendix A.3. Additionally, a decay factor can be incorporated to ensure effectiveness in long trajectories, as in the typical Bellman Equation.

\begin{algorithm}[ht]
\caption{The pseudo-code of IFT}
    \begin{algorithmic}[1]
        \STATE \textbf{Input:} \\
        Initial instruction $\rho_0$, Ground truth $s^*$ with $N$ tokens: $s^*[1], \ldots, s^*[N]$ \\
        ~~ \\
        
        \STATE \textbf{Step 1: Inference One Step Ahead}
        \FOR{$t$ in $[1,N]$}
            \STATE Predict the probability distribution of the $t$-th token: $P_t' = \mathcal{\pi_\theta}(s^*[0:t-1])$
            \STATE Sample tokens: $s^\theta[t] = \arg\max P_t'$
        \ENDFOR \\
        ~~ \\
        
        \STATE \textbf{Step 2: Intuitive Preference Estimation}
        \STATE Encode $s^*$ and $s^\theta$ using Embedding Layer $E$
        \STATE Compute the fused embedding: \\
        $e = (1-\lambda)E(s^*) + \lambda E(s^\theta)$
        \FOR{$t$ in $[1,N]$}
            \STATE Predict the probability distribution of the $t$-th token: $P_t'' = (\mathcal{\pi_\theta}/E)(e[0:t-1])$
            \STATE Compute token loss: $\mathcal{L}_t = \log(P_t'', s^*[t])$
        \ENDFOR \\
        ~~ \\
        
        \STATE \textbf{Step 3: Dynamic Relation Propagation}
        \FOR{$t$ in $[1,N]$}
            \STATE Compute the cumsum weight similar to Bellman Equation: $w_t = \sum\limits_{i=t}^N \alpha^{N-t}\mathcal{L}_i$
        \ENDFOR
        \\
        ~~ \\
        \STATE \textbf{Output:} Final loss $\mathcal{L}_{\text{IFT}} = w\cdot\mathcal{L}$
    \end{algorithmic}
\end{algorithm}

\begin{table*}[ht]
    \centering
    \begin{adjustbox}{width=\textwidth}
    \begin{tabular}{l|cccccc|c}
        \toprule
        \textbf{Method} & \textbf{ARC} & \textbf{ARC-Gen} & \textbf{MMLU} & \textbf{TruthfulQA} & \textbf{WinoGrande} & \textbf{GSM8K} & \textbf{Avg.} \\
        
        \midrule
        \midrule
        Mistral-7B
        & 53.07 & 73.04 & 59.14 & 45.29 & 77.58 & 38.89 & 54.79\\
        \midrule
        \multicolumn{8}{l}{ \quad \quad \textbf{\textit{fine-tuning with UltraFeedback-60k}}} \\
        \midrule
        + SFT 
        & 56.49 & 74.00 & 60.44 & 55.57 & 77.90 & 42.84 & 58.65\\
        
        + DPO 
        & \textbf{61.86} & 73.54 & \textbf{61.02} & 47.98 & 76.64 & 43.89 & 58.28\\

        + TDPO
        & 56.06 & 73.72 & 60.23 & 43.94 & 77.03 & 41.70 & 55.79\\
        
        + ORPO
        & 56.66 & 73.98 & 60.57 & 51.77 & 77.19 & 42.30 & 57.70\\

        + SimPO
        & 59.90 & 73.55 & 52.61 & 47.25 & 78.30 & 37.53 & 55.15\\
        
        + IFT
        & 56.74 & \textbf{74.15} & 60.49 & \textbf{57.65} & \textbf{78.45} & \textbf{44.73} & \textbf{59.61}\\
        \midrule
        \textcolor{gray}{Mistral-ORPO-$\alpha$}
        & \textcolor{gray}{57.25} & \textcolor{gray}{73.72} & \textcolor{gray}{58.74} & \textcolor{gray}{60.59} & \textcolor{gray}{73.72} & \textcolor{gray}{46.78} & \textcolor{gray}{59.41} \\
        \midrule
        \multicolumn{8}{l}{ \quad \quad \textbf{\textit{fine-tuning with Ultrachat-200k + UltraFeedback-60k sequentially}}} \\
        \midrule
        + SFT
        & 57.68 & 72.87 & 58.25 & 45.78 & 77.19 & 40.94 & 55.97\\
        
        + SFT + SFT
        & 58.10 & 72.61 & 58.40 & 48.59 & 76.80 & 43.06 & 56.99 \\
        
        + SFT + DPO
        & \textbf{63.91} & \textbf{73.98} & \textbf{59.75} & 46.39 & 76.06 & 41.47 & 57.52 \\

        + SFT + TDPO
        & 59.13 & 73.72 & 58.92 & 46.63 & 76.32 & \textbf{44.58} & 57.12\\

        + SFT + ORPO
        & 58.45 & 73.21 & 58.80 & 50.31 & 76.45 & 42.76 & 57.35 \\

        + SFT + SimPO
        & 60.83 & 73.63 & 59.01 & 49.45 & 76.95 & 38.44 & 56.94\\
        
        + SFT + IFT
        & 58.36 & 73.38 & 58.45 & \textbf{52.39} & \textbf{78.06} & 43.82 & \textbf{58.22} \\
        
        \midrule
        \textcolor{gray}{Zephyr-7B-$\beta$}
        & \textcolor{gray}{67.41} & \textcolor{gray}{72.61} & \textcolor{gray}{58.74} & \textcolor{gray}{53.37} & \textcolor{gray}{74.11} & \textcolor{gray}{33.89} & \textcolor{gray}{57.50} \\
        
        \bottomrule
        \end{tabular}
    \end{adjustbox}
    \caption{Evaluation on Open-LLM Leaderboard with chat template. When fine-tuning with the same recipe, IFT achieves the highest average score across all methods. Directly conducting alignment using IFT showcases the best performance in all recipes with the least data and computation.}
    \label{tab:llmleaderboard_chat}
\end{table*}

\begin{table*}[ht]
    \centering
    \begin{adjustbox}{width=\textwidth}
    \begin{tabular}{l|ccc|cccc|c}
        \toprule
        \multirow{2}{*}{\textbf{Method}} & \multirow{2}{*}{\textbf{Reference}} & \multicolumn{2}{c}{\textbf{Data}} & \multicolumn{2}{c}{\textbf{Alpaca-Eval}} & \multicolumn{2}{c}{\textbf{Alpaca-Eval-2}} & \textbf{TL;DR}\\
        & & pairwise & volume & win-rate & lc win-rate & win-rate & lc win-rate & win-rate\\
        
        \midrule
        \midrule
        Mistral-7B 
        & -- & -- & 120k
        & 24.72 & 11.57 & 1.25 & 0.35 & 92.03\\
        \midrule
        \multicolumn{6}{l}{ \quad \quad \textbf{\textit{fine-tuning with UltraFeedback-60k}}} \\
        \midrule
        + SFT
        & \XSolidBrush & \XSolidBrush & 120k
        & 82.56 & \underline{78.32} & 7.09 & 8.67 & 84.22 \\
        
        + DPO 
        & \Checkmark & \Checkmark & 120k
        & 74.00 & 73.12 & 9.73 & 8.58 & 77.25\\

        + TDPO
        & \Checkmark & \Checkmark & 120k
        & 65.74 & 51.41 & 4.99 & 3.47 & 70.82\\
        
        + ORPO
        & \XSolidBrush & \Checkmark & 120k
        & \underline{85.14} & 76.60 & 8.82 & 12.34 & \underline{89.24}\\

        + SimPO
        & \XSolidBrush & \Checkmark & 120k
        & 83.08 & 64.30 & \textbf{24.47} & \textbf{20.31} & 59.13 \\
        
        + IFT
        & \XSolidBrush & \XSolidBrush & 120k
        & \textbf{85.18} & \textbf{78.78} & \underline{9.95} & \underline{13.27} & \textbf{92.63}\\
        
        \midrule
        \textcolor{gray}{Mistral-ORPO-$\alpha$}
        & \textcolor{gray}{\XSolidBrush} & \textcolor{gray}{\Checkmark} & \textcolor{gray}{120k}
        & \textcolor{gray}{87.92} & \textcolor{gray}
        {--} & \textcolor{gray}{--} & \textcolor{gray}{11.33} & \textcolor{gray}{--}\\
        
        \midrule
        \multicolumn{6}{l}{ \quad \quad \textbf{\textit{fine-tuning with UltraChat-200k + UltraFeedback-60k sequentially}}} \\
        \midrule
        + SFT
        & \XSolidBrush & \XSolidBrush & 200k
        & 86.69 & 77.96 & 4.08 & 6.43 & 98.11\\
        
        + SFT + SFT
        & \XSolidBrush & \XSolidBrush & 260k
        & 86.34 & 76.98 & 4.55 & 7.14 & 97.79\\
        
        + SFT + DPO 
        & \Checkmark & \Checkmark & 320k
        & \textbf{91.62} & \textbf{81.54} & 10.08 & 13.72 & \textbf{99.18}\\

        + SFT + TDPO
        & \Checkmark & \Checkmark & 320k
        & \underline{89.80} & 76.44 & 9.25 & 14.15 & \underline{98.89} \\

        + SFT + ORPO
        & \XSolidBrush & \Checkmark & 320k
        & 86.26 & 79.67 & 7.40 & 12.27 & 97.92\\

        + SFT + SimPO
        & \XSolidBrush & \Checkmark & 320k
        & 88.79 & 68.88 & \textbf{19.62} & \textbf{23.94} & 98.23 \\
        
        + SFT + IFT
        & \XSolidBrush & \XSolidBrush & 260k
        & 88.37 & \underline{81.29} & \underline{10.26} & \underline{14.34} & 98.57 \\
        
        \midrule
        \textcolor{gray}{Zephyr-7B-$\beta$}
        & \textcolor{gray}{\Checkmark} & \textcolor{gray}{\Checkmark} & \textcolor{gray}{320k}
        & \textcolor{gray}{90.60} & \textcolor{gray}
        {--} & \textcolor{gray}{--} & \textcolor{gray}{10.99} & \textcolor{gray}{--}\\
        
        \bottomrule
        \end{tabular}
    \end{adjustbox}
    \caption{Evaluation on LLM-based Benchmarks. IFT secures top two rankings in nearly all tasks, including conversation and summarization. When fine-tuned on limited data from UltraFeedback, IFT demonstrates a significant lead in TL;DR.}
    \label{tab:llm}
\end{table*}

\section{Experiments}
\label{sec:experiments}
We conduct experiments mainly on NLP setting. Considering the absence of an optimal policy of human language generation, we also utilize the Frozen Lake environment for further validation.

\subsection{Settings for NLP}
\textbf{Datasets.} Our main experiments use UltraChat-200k \citep{ding2023enhancing} as the single-target dataset and UltraFeedback-60k \citep{cui2023ultrafeedback} as the pairwise preference dataset. We also include a variant of UltraFeedback-60k introduced by \citet{meng2024simpo}, which is sampled from Gemma2 and LLaMA3 and labeled with preferences using ArmoRM.

\textbf{Models.} Our main experiments are conducted on Mistral-7B-v0.1 \citep{jiang2023mistral} and Mistral-7B-sft-beta \citep{tunstall2023zephyr}, with the latter one fine-tuned from the former using UltraChat-200k. We also consider models with different architectures and parameter scales, including Gemma-2B \citep{team2024gemma} and LLaMA3-8B \citep{grattafiori2024llama}.

\textbf{Scenarios.} 
We consider two different training scenarios, one using Preference Optimization exclusively, and the other employing sequential recipe of SFT and Preference Optimization. In the first scenario, alignment is conducted directly from base model Mistral-7B-v0.1 using UltraFeedback. In order to ensure balanced data volume between different method, we randomly sample 60k data from UltraChat as supplementary for SFT and IFT, for only the target data are utilized in these two methods. The second scenario is commonly seen, where SFT and Preference Optimization is employed sequentially. For this scenario, we use Mistral-7B-sft-beta as start-point, which has been fine-tuned with UltraChat using SFT. Then we fine-tune it further with UltraFeedback using Preference Optimization.

\textbf{Baselines.} SFT and DPO \citep{rafailov2024direct} are our main baselines, and we exclude PPO due to computational limitations. We also incorporate three improved versions of DPO: TDPO \citep{zeng2024token}, ORPO \citep{hong2024reference}, and SimPO \citep{meng2024simpo}. 
TDPO transformers the DPO loss into token-level to make its objective closer to SFT. 
SimPO adds on a length-normalization term to replace the regularization from reference model.
ORPO adds the SFT loss and a DPO-like loss together, achieving alignment directly without SFT and reference model. 
In addition to reproducing the algorithms mentioned above, we also consider Zephyr-7B-beta \citep{tunstall2023zephyr} and Mistral-ORPO-alpha \citep{hong2024reference}, two open-source checkpoints that utilize sequential and direct recipes respectively. Both of them used start-point models and datasets similar to ours.

\textbf{Benchmarks.} We consider two types of benchmarks. One is from the widely used Open-LLM LeaderBoard, which contains ARC-Challenge(25-shot) \citep{clark2018think}, MMLU(5-shot) \citep{chung2024scaling}, TruthfulQA(0-shot) \citep{lin2021truthfulqa}, WinoGrande(5-shot) \citep{sakaguchi2021winogrande}, and GSM8K(5-shot) \citep{cobbe2021training}. 
The other is LM-based evaluation, including TL;DR \citep{volske2017tl}, Alpaca-Eval, and Alpaca-Eval-2 \citep{dubois2024alpacafarm}. As for TL;DR, we keep the same setting as \citep{rafailov2024direct}, using GPT-4 to judge the win-rate between model's generation and ground truth answer. We utilize chat template for all benchmarks to obtain a more accurate evaluation for chat models.

\subsection{Main Results in NLP Tasks}

\textbf{Effectiveness on Sequential Recipe.} 
In this scenario, IFT demonstrates good performance across benchmarks having standard answers or not (See Table \ref{tab:llmleaderboard_chat} and \ref{tab:llm} for details). 
On Open-LLM Leaderboard, IFT showcases the best average capabilities across all tasks, excelling particularly in tasks requiring generation, reasoning and fact-following abilities, such as TruthfulQA and GSM8K. 
However, IFT has a relatively large gap between DPO in multi-choice tasks like ARC-Challenge and MMLU.
When evaluated for conversation and summarization judged by GPT-4, IFT's performance is comparable to that of the chosen baselines. Remarkably, IFT achieves these results using the least amount of data and computational resources among all the methods tested.

\textbf{Effectiveness of Preference Optimization Alone.} IFT not only maintains the performance advantages compared with other baselines in this setting. But also, IFT performs comparably or even superiorly to many method in sequential recipe (See Table \ref{tab:llmleaderboard_chat}, \ref{tab:llm}, and Appendix \ref{appendix:experiments} for details). While DPO, SimPO and TDPO tend to fail under this setting, ORPO remains competitive in its open-source model. However, when constrained in the same experiment setting, the performance of ORPO becomes worse than IFT. Additionally, the reliance on preference data makes ORPO more costly in terms of negative sampling, preference labeling, and GPU memory consumption. Consequently, IFT stands out as a more efficient and cost-effective alternative in this context.

\textbf{Multi-Choice vs. Generation.} 
IFT performs better on generation tasks but struggles with multi-choice, whereas DPO exhibits the opposite performance. This may due to differences in evaluation metrics and training objectives \citep{zheng2023large, plaut2024softmax, tsvilodub2024predictions}. Multi-choice tasks evaluate log-likelihood for entire answers, while generation tasks require token-by-token construction for causality and reasoning. DPO aligns the mapping between instructions and complete answers, while IFT emphasizes token-level causal relationships. As a result, DPO tends to excel in multi-choice tasks, while IFT performs better in token-by-token exploration tasks. In an ARC-Challenge adaptation to generation tasks, IFT demonstrates superiority without changing the benchmark's distribution. Overall, IFT showcases its balanced performance across diverse tasks and achieving the highest average score.

\textbf{Objective Trade-off between SFT and Preference Optimization.}
Traditional Preference Optimization methods deliver excellent alignment performance, particularly in enhancing the instruction-following ability of language models, as showed in Table \ref{tab:llm}. However, fitting the different objectives of SFT and Preference Optimization involves trade-offs \citep{tunstall2023zephyr}. Even slight overfitting on SFT may result in reduced effectiveness of Preference Optimization. This phenomenon is also observed in Table \ref{tab:llmleaderboard_chat}, where the models trained by sequential recipe of SFT and other Preference Optimization methods showcase obvious inferior results on Open-LLM Leaderboard even worse than SFT alone. Avoiding this trade-off, ORPO and IFT can achieve better and more stable performance by directly conducting alignment on the base model.

\textbf{Efficiency and Scaling Potential of IFT.}
Although IFT achieves comparable or superior performance to other methods, it also boasts high efficiency in many aspects.
IFT does not require a reference model, which conserves GPU memory and computational resources. 
Most importantly, IFT and SFT are the only methods that conduct alignment without preference data, offering significant benefits as follows.
Firstly, this characteristic eliminates the need for synchronous storage and computation of pairwise data on the GPU, thereby reducing memory consumption and training duration. Secondly, negative sampling from models and human preference-labeling are no longer necessary, eliminating the highest cost associated with alignment, which has been a discarded but fundamental challenge in research so far. Furthermore, using only the target answer brings the potential for scaling in alignment or even in pre-training.

\subsection{Further Validation in Frozen-Lake Environment}

\begin{figure}[t]
    \centering
    \includegraphics[width=\columnwidth]{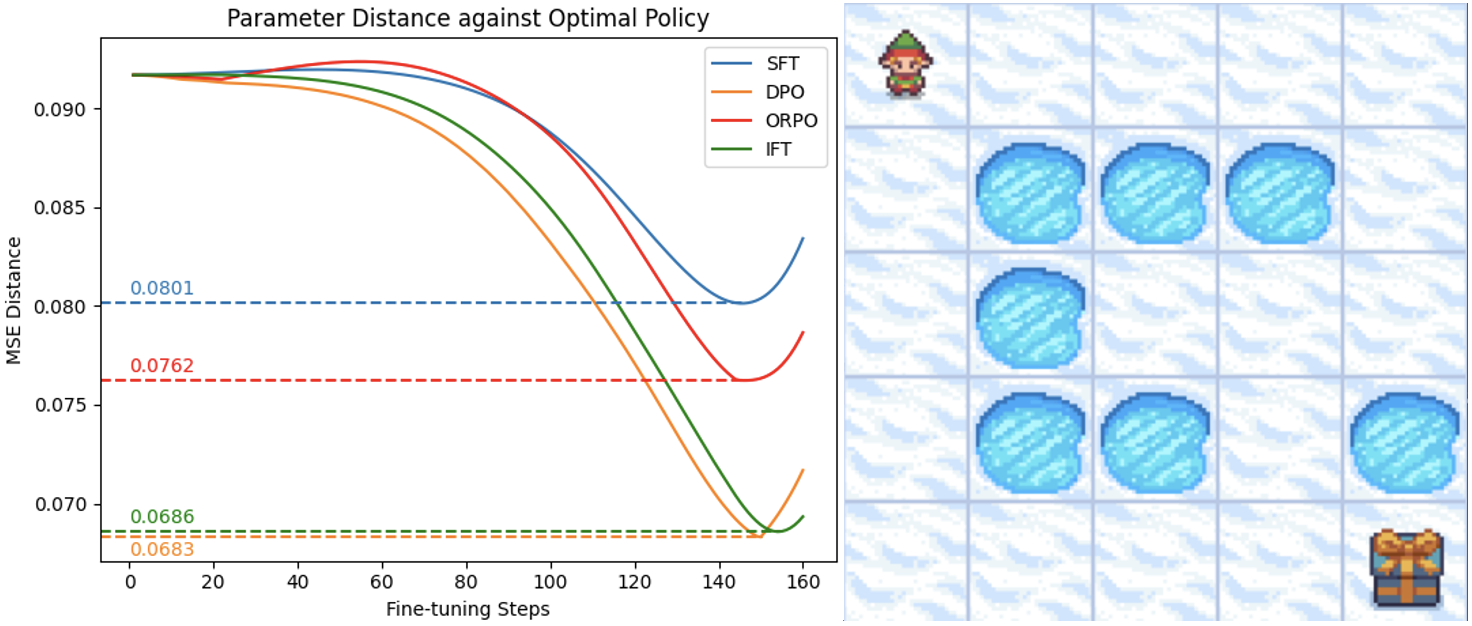}
    \caption{The Frozen Lake Game. Considering the MSE distance between transition matrices of the trained and optimal policy, IFT performs much better than SFT and ORPO, but slightly worse than DPO.}
    \label{fig:frozen-lake}
\end{figure}

As scores on Open-LLM Leaderboard only partially reflect models' performance, and GPT-4 inadequately models human language generation, further comparison to a truly optimal policy is necessary. Given the difficulty of obtaining an optimal policy representing human language, we validate our algorithm in a simplified setting called Frozen Lake \citep{frozenlake}. In this environment, an agent attempts to find a gift on a nearly frozen lake with several holes, terminating the game upon finding the gift or falling into a hole. The limited number of states and actions in this game allows the optimal policy to be easily derived using classical RL methods.

To simulate parameterized policy alignment, we employ a two-layer fully connected neural network and design the environment with one optimal and one sub-optimal trajectory. The optimal parameterized policy is trained using the previously obtained optimal state-action transition matrix, and various fine-tuning methods from LMs are compared. We evaluate performance by measuring the MSE distance between the transition matrices of the optimal and trained policy. We didn't count in TDPO and SimPO, as their objectives are similar as DPO in Frozen Lake Game.

In this setting, IFT achieves a significantly better policy than SFT and ORPO, although it performs slightly worse than DPO. This is partly because, in terms of comparing how closely the explored grid aligns with the agent's preference, the order is DPO $>$ IFT $>$ ORPO $>$ SFT. Although ORPO also considers the negative trajectories sampled from policy, its direct incorporation of SFT loss with a fusion coefficient deviates its preference estimation, partially diminishing its effectiveness. Additionally, DPO, ORPO and IFT explore more grids than SFT, which helps the agent develop a better understanding of the environment. 

\section{Related Work}
\label{sec:related_work}
\textbf{Classical Reinforcement learning (RL)} has demonstrated strong performance in various sequential decision-making and optimal control domains, including robotics \citep{levine2018learning}, computer games \citep{vinyals2019grandmaster} and others \citep{guan2021direct}. There are two main categories of RL algorithms: value-based and policy-based, depending on whether they learn a parameterized policy. Value-based RL aims to fit an value function defined by Bellman Equation, containing methods such as Monte-Carlo (MC) Learning \citep{lazaric2007reinforcement} and Temporal Difference Learning \citep{sutton1988learning, seijen2014true}. However, value-based methods struggle in continuous or large discrete space for its greedy objective. Thus, policy-based methods were introduced to model the decision-making process using a parameterized policy. As one of its best-known algorithms, Proximal Policy Optimization (PPO) \citep{schulman2017proximal} is widely used in various domains, including Natural Language Processing (NLP).

\textbf{Alignment for LMs} has emerged as a crucial task these years, which adjusts the LMs' generation distribution in line with human preferences \citep{bradley1952rank, ziegler2019fine, ouyang2022training, lee2023rlaif}. 
While PPO remains the primary algorithm for alignment, its high demands for computation and memory hinders its broader use. Consequently, many improved methods have been proposed \citep{dong2023raft, yuan2023rrhf, zhao2023slic}.
Among them, DPO \citep{rafailov2024direct} unifies reward modeling and policy optimization by utilizing a loss function derived from PPO, training a single model to serve as both a policy model and a reward model.
Without sacrificing performance, DPO decrease the costly consumption of PPO through directly value iteration similar to a preference-based format of MC instead of TD. However, it still relies on an expensive preference-labeling process and requires an SFT-based warm-up stage, which may introduce trade-offs when aligning the objectives of SFT and Preference Optimization.

\textbf{Improved Versions of DPO} come out one after another. 
Efforts such as \citep{liu2023makes, khaki2024rs, yin2024relative, guo2024controllable, bansal2024comparing, liu2024extensive} try to enhance the contrastive learning by utilizing better ranking strategies, more informative data, or more number of negative samples.
Except for using offline data, \citep{liu2023statistical, yuan2024self, guo2024direct, calandriello2024human, chen2024self, mitra2024orca} focus on online sampling and automated label/reward collection, reducing the manual cost required for alignment.
Methods like \citep{ethayarajh2024kto, hong2024reference} aim to reduce DPO's dependency on SFT warm-up by transforming its loss functions and data format into a SFT manner. These algorithms handle positive and negative samples using SFT objective and Unlikelihood Training \citep{welleck2019neural}, respectively. Recently, \citep{zeng2024token, meng2024simpo} improved the integration of the SFT and DPO by introducing various regularization terms. These terms prevent the policy model from overfitting DPO objective and deviating from SFT objective.
However, the actual volume of training data is not decreased in these methods. Also, GPU-memory-consuming pair-wise data is still required, while the need for a reference model and preference-labeling for the entire answer trajectory is only eliminated in limited cases.

\section{Conclusion}
\label{sec:conclusion}
In this paper, we first interpret SFT and typical Preference Optimization methods into a unified framework using Preference Estimation and Transition Optimization. Through this modeling, we found the biased prior used in SFT is one of the main reasons why SFT performs worse than other Preference Optimization methods. Then, we introduce an efficient and effective method called Intuitive Fine-Tuning (IFT), which achieves alignment directly from the base model using non-preference-labeled data. Finally, experiments on widely used NLP benchmarks and Frozen Lake environment demonstrate the competitive performance of IFT.

\section{Limitations} 
Our validation of IFT is limited to the fine-tuning setting, where data volume is constrained, leaving the scalability of IFT unexplored.

\section*{Acknowledgments}
This work is supported by the National Science and Technology Major Project (2023ZD0121403), and the Beijing Natural Science Foundation (IS23059). We further extend our gratitude to Yue Yu, Yihao Liu, Che Jiang, Xuekai Zhu, Jingkun Yang, Xuanqi Dong, Hong Liu, and Chushu Zhou for their insightful discussions with us.

\bibliography{acl}

\newpage
\appendix

\section{Theoretical Details}
\label{appendix:theory}
\subsection{MDP In LMs}
\label{appendix:mdp_in_lms}
$\mathcal{M}=(S, A, \mathcal{T}, r, \rho_0)$:
\begin{itemize}
    \item $A$, the concrete action space, consisting of $N_A$ vocabularies as defined by the tokenizer.
    \item $S$, the concrete state space, comprising $N_S=(N_A)^N$ elements related to sequence length $N$. Each state represents a ordered permutation of vocabularies.
    \item $\rho_0$, the initial state of each generation, typically refers to the given instruction;
    \item $\mathcal{T} \in R^{N_S \times N_A}$, the state-action transition matrix of a given policy, indicating the probability of generating each token given different states;
    \item $r$, the reward assigned to a particular state-action pair.
\end{itemize}

\subsection{Loss Function of IFT}
\label{appendix:loss_function_derivation}
The disparities of transition between model and human can be formalized as follows:

\begin{equation}
    \begin{aligned}
        \mathcal{L}(\hat{T_\theta};T^*) &= \mathbb{E}_{\rho_0\sim\mathcal{D}}\mathbb{E}_{s_n^*\sim\mathcal{S}_{\rho_0}^*}\\
        &\left[-\sum\limits_{n=0}^N\log\frac{\hat{T_\theta}(s_n^*, \rho_0)}{T^*(s_n^*, \rho_0)}\right]
    \end{aligned}
\end{equation}
We make the same hypothesis as SFT that the optimization objective of each target intermediate state has a probability equal to 1, so that $\forall n \in [0, N], \rho_0\in\mathcal{D}, s_n^*\in\mathcal{S}_{\rho_0}^*$: 
\begin{equation}
    T^*(s_n^*, \rho_0) = 1 = T^*(s_N^*, \rho_0)
\end{equation}
Thus, the objective of IFT can be represented directly by the following loss function:
\begin{equation}
    \begin{aligned}
        \mathcal{L}(\hat{T_\theta}) &= \mathbb{E}_{\rho_0\sim\mathcal{D}} \mathbb{E}_{s_i^*\sim\mathcal{S}_{\rho_0}^*} \\
        &\left[ -\sum_{n=0}^N \log \mathcal{T}_\theta \left( \pi^* \left( \delta^* (s_i^*) \right), \delta_\theta (s_i^*) \right) \right]
    \end{aligned}
\end{equation}
As the optimal policy enjoys the optimal transition:
\begin{equation}
    s_i^* = [s_{i-1}^*, a_i^*] = [s_{i-1}^*, \pi^*(s_{i-1}^*)] = \Pi^*(s_{i-1}^*)
\end{equation}
Therefore, the disturbed optimal state keeps similar with the original optimal state:
\begin{equation}
    \delta^*(s_i^*)=(1-\lambda)s_i^* + \lambda\Pi^*(s_{i-1}^*) = s_i^*
\end{equation}
Then, the final loss function can be presented as:
\begin{equation}
    \begin{aligned}
        \mathcal{L}(\mathcal{T}_\theta,\delta_\theta) &= \mathbb{E}_{\rho_0\sim\mathcal{D}}\mathbb{E}_{s_i^*\sim\mathcal{S}_{\rho_0}^*}\\
        &\left[-\sum\limits_{n=0}^N\log\mathcal{T}_\theta(a_i^*, \delta_\theta(s_i^*))\right]
    \end{aligned}
\end{equation}

\subsection{Proof for Bellman Equation}
\label{appendix:bellman_equation_derivation}
Considering only one sampled state $s_n^*$ constrained by $\rho_0$ in the datasets, we have:
\begin{equation}
    \begin{aligned}
        &\exp\big(-\mathcal{L}(\hat{T_\theta}(s_n^*, \rho_0))\big) \\
        &=\mathcal{T}_\theta(a_n^*, \delta_\theta(s_n^*))\bigg(\sum\limits_{n+1}^N\mathcal{T}_\theta(a_i^*, \delta_\theta(s_i^*))\bigg) \\
        &=\max\limits_{a}\bigg[\mathcal{T}_\theta(a, s_n^*)\big(r+\gamma V(\hat{s^\theta_{n+1}})\big)\bigg] \\
        &= V_\theta(\hat{s_n^\theta})
    \end{aligned}
\end{equation}
where $r=(1-\gamma) V(\hat{s^\theta_{n+1}})$. This reward function implicitly accounts for the influence of the current prediction on future generations.

\subsection{Reformulation of Typical Methods}
\label{appendix:reformulation}
We reformulate the loss function of some methods using the disparities of transition matrices as:

\textbf{SFT}
\begin{equation}
    \mathcal{L}_{\text{SFT}} 
    = \mathbb{E}_{\rho_0\sim\mathcal{D}}\mathbb{E}_{s_i^*\sim\mathcal{S}_{\rho_0}^*}
    \left[-\sum\limits_{i=0}^N\log\mathcal{T}_\theta(\pi^*(s_i^*), s_i^*)\right]
\end{equation}
where the human's preference is unbiasedly estimated, but the model's preference is inaccurately represented by $s_i^*$.

\textbf{PPO}
\begin{equation}
    \mathcal{L}_{\text{PPO}}
    = \mathbb{E}_{\rho_0\sim\mathcal{D}}\mathbb{E}_{s_i^*\sim\mathcal{S}_{\rho_0}^*}
    \left[-\sum\limits_{i=0}^N \mathcal{R}(\pi_\theta(s_i^\theta),s_i^\theta)\right]
\end{equation}
where $\mathcal{R} \in (-\infty, 0]$ denotes the degree of closeness between human preferences and the state-action pairs chosen by model. The reward and loss will be zero only if the state-action pairs perfectly align with human preferences. Thus, PPO implicitly models the human policy $\pi^*$ through reward modeling, which can be formulated as follows:
\begin{equation}
    \mathcal{R}=\pi_\mathcal{R} \leftarrow \min\limits_\pi \mathcal{L_R}
\end{equation}
\begin{equation}
    \begin{aligned}
        \mathcal{L_R} &= \mathbb{E}_{\rho_0\sim\mathcal{D}}\mathbb{E}_{s_i^+\sim\mathcal{S}_{\rho_0}^+, s_i^-\sim\mathcal{S}_{\rho_0}^-} \\
        &\left[-\log\sigma \bigg(\sum\limits_{i=0}^N\log\mathcal{T_R}(\pi^+(s_i^+)|s_i^+) \right.\\
        &\left. \quad\quad\quad\quad\quad - \sum\limits_{i=0}^N\log\mathcal{T_R}(\pi^-(s_i^-)|s_i^-)\bigg)\right]
    \end{aligned}
\end{equation}

\textbf{DPO-Online}
\begin{equation}
    \begin{aligned}
        \mathcal{L}_{\text{DPO}} &= \mathbb{E}_{\rho_0\sim\mathcal{D}}\mathbb{E}_{s_i^*\sim\mathcal{S}_{\rho_0}^*, s_i^\theta\sim\mathcal{S}_{\rho_0}^\theta} \\
        & \left[-\log\sigma\bigg(\sum\limits_{i=0}^N\log\mathcal{T}_\theta(\pi^*(s_i^*), s_i^*) \right.\\
        & \left. \quad\quad\quad\quad\quad -\sum\limits_{i=0}^N\log\mathcal{T}_\theta(\pi_\theta(s_i^\theta),s_i^\theta)\bigg)\right]
    \end{aligned}
\end{equation}
Ideally, this loss function increases the probabilities of state-action pairs preferred by humans and decreases the probabilities of those chosen by the model. It unbiasedly estimate both the human's and model's preference.

\textbf{DPO-Offline}
\begin{equation}
    \begin{aligned}
        \mathcal{L}_{\text{DPO}} &= \mathbb{E}_{\rho_0\sim\mathcal{D}}\mathbb{E}_{s_i^+\sim\mathcal{S}_{\rho_0}^+, s_i^-\sim\mathcal{S}_{\rho_0}^-} \\
        & \left[-\log\sigma\bigg(\sum\limits_{i=0}^N\log\mathcal{T}_\theta(\pi^+(s_i^+), s_i^+) \right.\\
        & \left. \quad\quad\quad\quad\quad -\sum\limits_{i=0}^N\log\mathcal{T}_\theta(\pi^-(s_i^-),s_i^-)\bigg)\right]
    \end{aligned}
\end{equation}
In the offline circumstance, the positive samples can still represent the human preference correctly, as $s^+$ is usually similar to $s^*$. However, this is not the case for negative samples.As training progresses, $s^-$ becomes more and more out-of-distributions compared to the model's preferred state $s^\theta$, leading to biased estimations.

\textbf{IFT}
\begin{equation}
    \begin{aligned}
        \mathcal{L}_{\text{IFT}}
        &= \mathbb{E}_{\rho_0\sim\mathcal{D}}\mathbb{E}_{s_i^*\sim\mathcal{S}_{\rho_0}^*}\\
        &\left[-\sum\limits_{n=0}^N\sum\limits_{i=n}^N\log\mathcal{T}_\theta(a_i^*, \delta_\theta(s_i^*))\right]
    \end{aligned}
\end{equation}
\begin{equation}
    \delta_\theta(s_i^*) = (1-\lambda)s_i^* + \lambda\pi_\theta(s_{i-1}^*)
\end{equation}
By using a model-based disturbance function, IFT constructs a residual connection in the temporal dimension, providing a better estimation for the model than SFT. Through this approach, IFT implicitly implements a Relation Propagation in the Transition Optimization stage, which considers the influence of current predictions on future outcomes. This propagation also reduces the influence of bias introduced by inaccurate estimations in earlier positions.

\section{Implementation Details}
\label{appendix:implementation}

\subsection{NLP Settings}

For the coefficient $\beta$ in DPO, TDPO, ORPO and SimPO, we use 0.1, 0.1, 0.25 and 2.0 respectively, as presented in their original papers. For the coefficient $\gamma/\beta$ ration in SimPO, we use 0.8 to keep the same setting in its original papers. For IFT, we choose 0.2 for $\lambda$ and incorporate a decay factor of 0.95 to fitting better with the Bellman Equation. We save checkpoints every 20k steps and select the results from the checkpoint with the best average score to demonstrate the performance of each method.

\begin{table}[htbp]
    \centering
    \tabcolsep=1cm
	\renewcommand\arraystretch{1.2}
    \begin{adjustbox}{width=\columnwidth}
        \begin{tabular}{cc}
            \toprule
            \textbf{Name} & \textbf{Value} \\
            \midrule
            epoch & 3 \\
            mini batch size & 8 \\
            gradient accumulation step & 64 \\
            warmup ratio & 0.1 \\
            scheduler & cosine \\
            learning rate & 5e-7 \\
            optimizer & RMSprop \\
            precision & bfloat16 \\
        \bottomrule
    \end{tabular}
    \end{adjustbox}
    \caption{Hyper-Parameters in NLP Setting}
    \label{tab:hyper}
\end{table}

We implement our main experiments on four NVIDIA A6000 GPUs. When using 60k single-target data, the entire training process for SFT and IFT takes approximately 20 hours, with each epoch lasting 7 hours. When using 60k pair-wise data, the training process for DPO and ORPO takes around 40 hours and 30 hours respectively, due to the differences in requirements for a reference model.

\subsection{Frozen Lake Setting}

We keep the similar hyper-parameters as in NLP setting for Frozen Lake game, running this environment on CPUs. Since our designed environment includes an optimal and a sub-optimal trajectory, we select the optimal trajectory as the target for SFT and IFT. For DPO and ORPO, the optimal and sub-optimal trajectories are used as positive and negative samples, respectively.

\section{More Experimental Results}
\label{appendix:experiments}

\begin{table*}[h]
    \centering
    \begin{adjustbox}{width=\textwidth}
    \begin{tabular}{l|cccccc|c}
        \toprule
        \textbf{Method} & \textbf{ARC} & \textbf{ARC-Gen} & \textbf{MMLU} & \textbf{TruthfulQA} & \textbf{WinoGrande} & \textbf{GSM8K} & \textbf{Avg.}\\
        \midrule
        \midrule
        Gemma-2B
        & 42.75 & 43.17 & 35.68 & 35.25 & 66.46 & 16.98 & 39.42\\
        \midrule
        \multicolumn{6}{l}{ \quad \quad \textbf{\textit{fine-tuning with Gemma2-UltraFeedback-armnorm-60k}}} \\
        \midrule
        + SFT 
        & \underline{42.06} & \textbf{42.75} & 34.30 & 41.49 & 64.88 & \underline{21.53} & 40.85\\
        
        + DPO 
        & 41.30 & 40.61 & 35.47 & 30.11 & 65.51 & 18.95 & 38.27\\

        + TDPO
        & 41.21 & 40.70 & 35.62 & 31.33 & 65.04 & 18.88 & 38.42\\
        
        + ORPO
        & 41.89 & 42.06 & \textbf{36.43} & \underline{41.98} & \underline{65.90} & 20.54 & \underline{41.35}\\

        + SimPO
        & 41.38 & 40.10 & 35.32 & 28.76 & 65.27 & 20.39 & 38.22\\
        
        + IFT
        & \textbf{42.49} & \underline{42.66} & \underline{35.77} & \textbf{45.41} & \textbf{66.14} & \textbf{22.14} & \textbf{42.39}\\

        \midrule
        \midrule
        LLaMA-3-8B
        & 49.40 & 73.89 & 62.17 & 46.63 & 76.80 & 50.26 & 57.05\\
        \midrule
        \multicolumn{6}{l}{ \quad \quad \textbf{\textit{fine-tuning with LLaMA3-UltraFeedback-armnorm-60k}}} \\
        \midrule
        + SFT 
        & 52.83 & 75.00 & \underline{63.24} & 50.42 & 76.95 & 51.09 & 58.91\\
        
        + DPO 
        & 51.19 & 74.23 & 62.21 & 36.35 & 76.24 & 51.25 & 55.45\\

        + TDPO
        & 51.37 & 74.31 & 62.50 & 39.66 & 76.50 & 51.73 & 56.36\\
        
        + ORPO
        & \underline{54.18} & \underline{74.98} & \textbf{63.46} & \underline{54.83} & \underline{77.06} & 51.18 & \underline{60.14}\\

        + SimPO
        & 53.92 & 74.48 & 62.76 & 38.07 & 76.56 & \textbf{51.93} & 56.57\\
        
        + IFT
        & \textbf{54.69} & \textbf{75.08} & 63.20 & \textbf{57.64} & \textbf{77.27} & \underline{51.78} & \textbf{60.92}\\
        \bottomrule
        \end{tabular}
    \end{adjustbox}
    \caption{Evaluation on Open-LLM Leaderboard when fine-tuning with UltraFeedback-60k.}
    \label{tab:llmleaderboard_chat-ultra-gemma}
\end{table*}

\begin{table*}[h]
    \centering
    \begin{adjustbox}{width=\textwidth}
    \begin{tabular}{l|ccc|cccc}
        \toprule
        \multirow{2}{*}{\textbf{Method}} & \multirow{2}{*}{\textbf{Reference}} & \multicolumn{2}{c}{\textbf{Data}}& \multicolumn{2}{c}{\textbf{Alpaca-Eval}} & \multicolumn{2}{c}{\textbf{Alpaca-Eval-2}} \\
        & & pairwise & volume & win-rate & lc win-rate & win-rate & lc win-rate \\
        \midrule
        \midrule
        Gemma-2B 
        & -- & -- & --
        & -- & -- & -- & -- \\
        \midrule
        \multicolumn{6}{l}{ \quad \quad \textbf{\textit{fine-tuning with UltraFeedback-60k}}} \\
        \midrule
        + SFT 
        & \XSolidBrush & \XSolidBrush & 120k
        & 36.53 & 30.28 & 0.99 & 0.57 \\
        
        + DPO 
        & \Checkmark & \Checkmark & 120k
        & 3.13 & 1.18 & 0.13 & 0.23 \\

        + TDPO
        & \Checkmark & \Checkmark & 120k
        & 2.14 & 0.70 & 0.25 & 0.10 \\
        
        + ORPO
        & \XSolidBrush & \Checkmark & 120k
        & \underline{36.62} & \underline{34.23} & 1.12 & 0.59 \\

        + SimPO
        & \XSolidBrush & \Checkmark & 120k
        & 4.48 & 2.42 & 0.13 & 0.15 \\
        
        + IFT
        & \XSolidBrush & \XSolidBrush & 120k
        & \textbf{36.74} & \textbf{39.33} & 1.61 & 1.23 \\

        \midrule
        \multicolumn{6}{l}{ \quad \quad \textbf{\textit{fine-tuning with Gemma2-UltraFeedback-armnorm-60k}}} \\
        \midrule
        + SFT 
        & \XSolidBrush & \XSolidBrush & 120k
        & 39.33 & 32.36 & 0.86 & 0.69 \\
        
        + DPO 
        & \Checkmark & \Checkmark & 120k
        & 2.83 & 0.81 & 0.00 & 0.00 \\

        + TDPO
        & \Checkmark & \Checkmark & 120k
        & 2.41 & 0.60 & 0.00 & 0.00 \\
        
        + ORPO
        & \XSolidBrush & \Checkmark & 120k
        & \underline{43.46} & \underline{34.19} & 2.06 & 1.21 \\

        + SimPO
        & \XSolidBrush & \Checkmark & 120k
        & 3.24 & 1.07 & 0.00 & 0.00 \\
        
        + IFT
        & \XSolidBrush & \XSolidBrush & 120k
        & \textbf{51.23} & \textbf{37.76} & 2.14 & 1.33 \\
        \bottomrule
        \end{tabular}
    \end{adjustbox}
    \caption{Evaluation on LLM-based Benchmarks when fine-tuning with UltraFeedback-60k.}
    \label{tab:llm-ultra-gemma}
\end{table*}

\begin{table*}[t]
    \centering
    \begin{adjustbox}{width=\textwidth}
    \begin{tabular}{l|cccccc|c}
        \toprule
        \textbf{Method} & \textbf{ARC} & \textbf{ARC-Gen} & \textbf{MMLU} & \textbf{TruthfulQA} & \textbf{WinoGrande} & \textbf{GSM8K} & \textbf{Avg.} \\
        
        \midrule
        \midrule
        Mistral-7B
        & 53.07 & 73.04 & 59.14 & 45.29 & 77.58 & 38.89 & 54.79\\
        + SFT 
        & 56.49 & 74.00 & 60.44 & 55.57 & 77.90 & 42.84 & 58.65\\
        
        + IFT
        & 56.74 & 74.15 & 60.49 & \textbf{57.65} & \textbf{78.45} & 44.73 & 59.61\\

        + IFT with noisy lambda
        & \textbf{61.60} & \textbf{76.53} & \textbf{61.11} & 57.03 & 77.43 & \textbf{45.64} & \textbf{60.56} \\
        
        \bottomrule
        \end{tabular}
    \end{adjustbox}
    \caption{Evaluation on Open-LLM Leaderboard when fine-tuning with UltraFeedback-60k.}
    \label{tab:abb}
\end{table*}

\end{document}